\pdfoutput=1
\documentclass[11pt]{article}
\usepackage[final]{acl}
\usepackage{times}
\usepackage{latexsym}
\usepackage[T1]{fontenc}
\usepackage[utf8]{inputenc}
\usepackage{microtype}
\usepackage{inconsolata}
\usepackage{graphicx}
\usepackage{float}
\usepackage{listings}
\title{SeQwen at the Financial Misinformation Detection Challenge Task: Sequential Learning for Claim Verification and Explanation Generation in Financial Domains} 
\author{Jebish Purbey \\
  Pulchowk Campus, IoE \\
  \texttt{jebishpurbey@gmail.com} \\
  \And Siddhant Gupta \\
  IIT Roorkee \\
  \texttt{siddhant\_g@me.iitr.ac.in} \\ 
  \And Nikhil Manali \\
 State University of New York, Buffalo\\
  \texttt{nmanali@buffalo.edu} \\ \\
  \AND Siddartha Pullakhandam * \\
  University of Wisconsin \\
  \texttt{pullakh2@uwm.edu} \\ \\
  \And Drishti Sharma * \\
  Cohere For AI Community\\
  \texttt{drishtishrma@gmail.com} \\ \\
  \And Ashay Srivastava * \\
  University of Maryland \\
  \texttt{ashays06@umd.edu} \\
  \AND Ram Mohan Rao Kadiyala  \\
  University of Maryland \\
  \texttt{rkadiyal@umd.edu} \\ }
\begin{document}
\maketitle
\begin{abstract}
This paper presents the system description of our entry for the COLING 2025 FMD challenge, focusing on misinformation detection in financial domains. We experimented with a combination of large language models, including Qwen, Mistral, and Gemma-2, and leveraged pre-processing and sequential learning for not only identifying fraudulent financial content but also generating coherent, and concise explanations that clarify the rationale behind the classifications.
Our approach achieved competitive results with an F1-score of \(0.8283\) for classification, and ROUGE-1 of \(0.7253\) for explanations. This work highlights the transformative potential of LLMs in financial applications, offering insights into their capabilities for combating misinformation and enhancing transparency while identifying areas for future improvement in robustness and domain adaptation.
\end{abstract}
\makeatletter
\def\blfootnote{\gdef\@thefnmark{}\@footnotetext}
\makeatother
\blfootnote{* equal contribution}
\section{Introduction}
Information is the backbone of the financial sector, supporting decision-making, market stability, risk management, regulatory compliance, and trust. However, the growth of digital media has increased the spread of financial misinformation. Misleading claims can influence markets and skew economic perceptions, posing serious hazards to institutions and investors. 
With the rise of large language models (LLMs), there is an opportunity to tackle this challenge effectively. LLMs have already demonstrated their potential in financial analysis \cite{shah-etal-2022-flue}, predictions \cite{wu2023bloomberggptlargelanguagemodel}, and decision-making \cite{xie2023pixiulargelanguagemodel}. In light of this, this paper focuses on our submission to the COLING 2025 Financial Minsinformation Detection (FMD) challenge, involving two key tasks: a three-way classification of financial claims backed by justifications for each classification. Our system enhances the capabilities of open-source LLMs for FMD by sequentially fine-tuning it to classify and generate explanations. We test a multitude of open-source models and select the best model for sequential learning. Our work contributes to developing specialized LLMs in financial domains for finer decision-making.
\begin{figure*}[ht]
    \centering
  \includegraphics[scale=0.47]{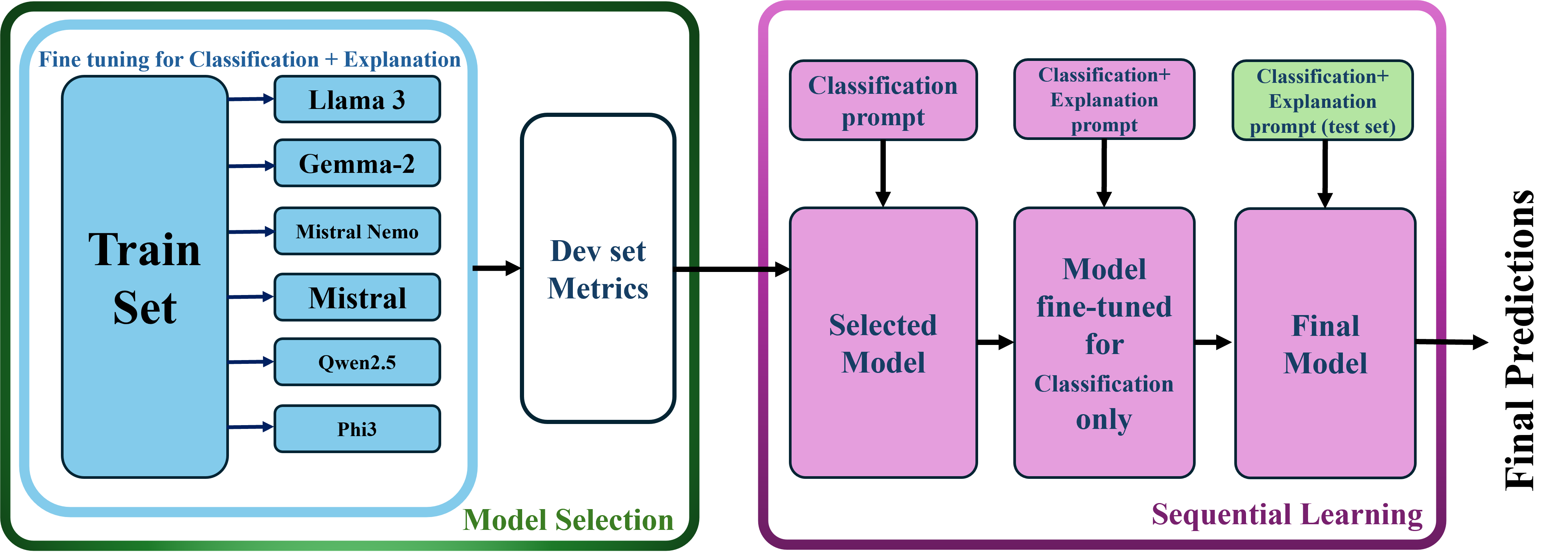}
  \caption{System design workflow. The development set is initially used to select the best-performing model, which is then fine-tuned on the train set using the sequential learning approach. The final model is then used for inference on the test set.}
  \label{fig:system}
\end{figure*}

\section{Dataset \& Task}
\begin{figure}
    \centering
    \includegraphics[width=0.9\linewidth]{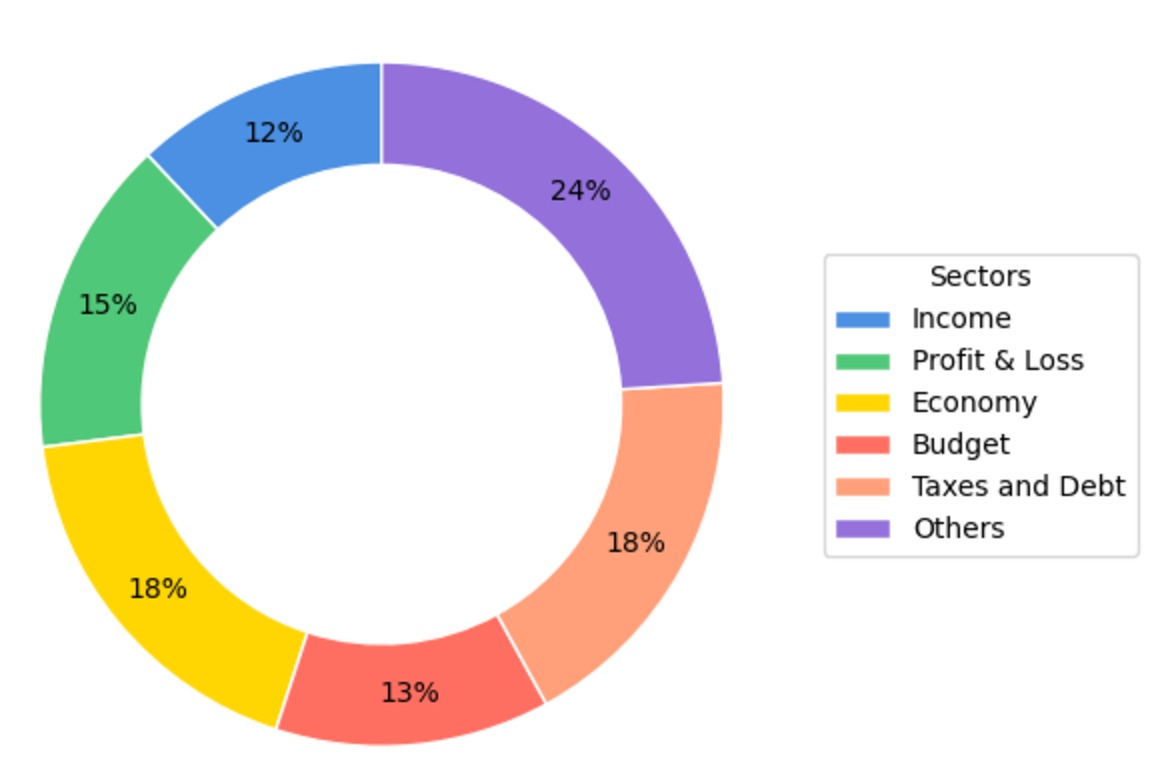}
    \caption{Distribution of financial claims across different sectors. Adapted from \cite{rangapur2024finfactbenchmarkdatasetmultimodal}.}
    \label{fig:dist}
\end{figure}

FMD challanege focuses on advancing LLM capabilities to detect financial misinformation while providing clear, evidence-based explanations for their decisions. Connecting claims with contextual information, these explanations aim to make the AI’s decisions more transparent, increasing trust and practicality for users, including investors and regulators. The task leverages the FIN-FACT \cite{rangapur2024finfactbenchmarkdatasetmultimodal} dataset which includes claims categorized as True, False, or Not Enough Information (NEI) across diverse sectors, including Income, Profit \& Loss, Economy, Budget, Taxes, and Debt, as visualized in Figure \ref{fig:dist}. The training set consists of 1953 samples with 1304 samples in the test set. For the purpose of model selection, the training set is split into train and dev sets, whose distributions are as shown in Table \ref{tab:dataset}.
\begin{table}[t]
\centering
\begin{tabular}{|l|c|c|}
\hline
\textbf{Class}     & \textbf{Train} & \multicolumn{1}{l|}{\textbf{Dev}} \\ \hline
\textbf{False (0)} & 696            & 196                               \\
\textbf{True (1)}  & 542            & 175                               \\
\textbf{NEI (2)}   & 262            & 82                                \\ \hline
\textbf{Total}     & 1500           & \multicolumn{1}{l|}{453}          \\ \hline
\end{tabular}
\caption{Class distribution for the train and dev set}
\label{tab:dataset}
\end{table}

\section{Methodology}
For the FMD challenge, we formulate the task as text generation and design the prompt to generate classification and explanations from the model simultaneously as in \cite{liu2024fmdllamafinancialmisinformationdetection}. Our main approach involves using sequential learning for the task, where we first fine-tune the LLM for classification only, followed by a second stage of fine-tuning for simultaneous classification and explanation generation, as shown in Figure \ref{fig:system}.\\ For the purpose of model selection, we fine-tune 5 open-source LLMs for the classification of financial claims. We then select the best-performing models and fine-tune them for joint classification and explanation generation. For evaluation, we use the micro F1 score for classification and ROUGE (1, 2, and L) \cite{lin-2004-rouge} for explanation generation as the performance metrics on the development set. The models fine-tuned under this approach include Qwen2.5 \cite{qwen2.5}, LLama3 8B \cite{dubey2024llama3herdmodels}, Mistral 7B \cite{jiang2023mistral7b}, Phi3 medium 4K Instruct \cite{abdin2024phi3technicalreporthighly}, and Gemma-2 9B \cite{gemmateam2024gemmaopenmodelsbased}. All the models were fine-tuned for 3 epochs with a learning rate of 2e-4, max sequence length of 1024, and total batch size of 16 for classification. For explanation generation, we fine-tuned the models for 5 epochs with all other hyperparameters same as the classification fine-tuning. Finally, we fine-tune the best-performing model in the sequential learning approach and compare the results with its single-stage training counterpart in the dev and test set.\\
All the fine-tuning of models were carried out using Unsloth with low-Rank Adaptation of Large Language Models (LoRA) \cite{hu2021loralowrankadaptationlarge}. The values for both the rank (\(r\)) and alpha (\(\alpha\)) were set to 16. For fine-tuning the model for classification only, we design the input prompt to include only labels. For simultaneous classification and explanation generation, we design the prompt to include both the label and evidence in the input. The difference between the two prompts is displayed in figure \ref{fig:prompts}. We utilize claims, justifications, labels, and evidence as our input for fine-tuning. We employed a preprocessing step where we appended some "claims" from the "justification" field, during the fine-tuning phase.
\lstset{
    basicstyle=\small\ttfamily,
    breaklines=true,
    breakatwhitespace=true,
    frame=single,
    columns=fullflexible
}
\begin{figure*}[!ht]
\centering
\begin{minipage}{0.45\textwidth}
\begin{lstlisting}
Below is an instruction that describes a task, paired with a claim and justification that provides further context. Write a response that appropriately completes the request.

### Instruction:
The goal is to classify the text as true/not_enough_info/false. Choose the correct category from these options and add an explanation after an empty line:
1: True
2: NEI
3: False

### Claim:
{claim}

### Justification:
{justification}

### Response:
{label}
\end{lstlisting}
\end{minipage}
\hfill
\begin{minipage}{0.45\textwidth}
\begin{lstlisting}
Below is an instruction that describes a task, paired with a claim and justification that provides further context. Write a response that appropriately completes the request.

### Instruction:
The goal is to classify the text as true/not_enough_info/false. Choose the correct category from these options and add an explanation after classification:
1: True
2: NEI
3: False
Your response must be in the following format:
Prediction: Your_Prediction Explanation: Your_Explanation

### Claim:
{claim}
### Justification:
{justification}
### Response:
Prediction: {label} Explanation: {expl}
\end{lstlisting}
\end{minipage}
\caption{Comparison of prompts used for classification and classification \& explanation generation.}
\label{fig:prompts}
\end{figure*}
\begin{table}[t]
\centering
\begin{tabular}{|l|c|}
\hline
\textbf{Model}            & \textbf{Micro F1} \\ \hline
Llama3 8B        & 0.8190            \\
Mistral 7B       & 0.8234            \\
Qwen2.5 7B       & \textbf{0.8455}   \\
Qwen2.5 32B      & 0.7947            \\
Phi 3 Medium     & 0.6733            \\
Gemma-2 9B       & 0.8035            \\ \hline
\end{tabular}
\caption{Performance on the dev set for classification}
\label{tab:class}
\end{table}
\begin{table*}
\centering
\resizebox{\textwidth}{!}{%
\begin{tabular}{|l|l|c|c|c|c|c|}
\hline
\textbf{Model} &
  \textbf{Description} &
  \multicolumn{1}{l|}{\textbf{Micro F1}} &
  \textbf{ROUGE-1} &
  \multicolumn{1}{l|}{\textbf{ROUGE-2}} &
  \multicolumn{1}{l|}{\textbf{ROUGE-L}} &
  \multicolumn{1}{l|}{\textbf{Overall Score}} \\ \hline 
Mistral 7B &
  \multicolumn{1}{l|}{\begin{tabular}[c]{@{}l@{}}Mistral 7B fine-tuned for classification and explanation\\ generation for a total of 5 epochs\end{tabular}} &
  \multicolumn{1}{c|}{0.7837} &
  0.6710 &
  0.6158 &
  0.6279 &
  0.7274 \\  
Qwen2.5 7B 5ep &
  \multicolumn{1}{l|}{\begin{tabular}[c]{@{}l@{}}Qwen2.5 7B fine-tuned for classification and explanation\\ generation for a total of 5 epochs\end{tabular}} &
  \multicolumn{1}{c|}{0.8322} &
  0.6710 &
  0.6133 &
  0.6333 &
  0.7516 \\
Qwen2.5 7B 8ep &
  \multicolumn{1}{l|}{\begin{tabular}[c]{@{}l@{}}Qwen2.5 7B fine-tuned for classification and explanation\\ generation for a total of 8 epochs\end{tabular}} &
  \multicolumn{1}{c|}{0.8234} &
   0.6871&
   0.6217&
   0.6447&
   0.7552\\
SeQwen &
  \multicolumn{1}{l|}{\begin{tabular}[c]{@{}l@{}}Qwen2.5 7B fine-tuned using sequential learning approach\\ for a total of 8 epochs (3 epochs of classification followed by\\ 5 epochs of classification + explanation generation)\end{tabular}} &
  \multicolumn{1}{c|}{\textbf{0.8366}} &
  \textbf{0.7170} &
  \textbf{0.6639} &
  \textbf{0.6772} &
  \textbf{0.7768} \\ \hline
\end{tabular}%
}
\caption{Performance on the dev set for Financial Misinformation Detection}
\label{tab:devFMD}
\end{table*}
\begin{table*}
\centering
\resizebox{\textwidth}{!}{%
\begin{tabular}{|l|l|c|c|c|c|c|}
\hline
\textbf{Model} &
  \textbf{Description} &
  \multicolumn{1}{l|}{\textbf{Micro F1}} &
  \textbf{ROUGE-1} &
  \multicolumn{1}{l|}{\textbf{ROUGE-2}} &
  \multicolumn{1}{l|}{\textbf{ROUGE-L}} &
  \multicolumn{1}{l|}{\textbf{Overall Score}} \\ \hline
Qwen2.5 7B 5ep &
  \multicolumn{1}{l|}{\begin{tabular}[c]{@{}l@{}}Qwen2.5 7B fine-tuned for classification and explanation\\ generation for a total of 5 epochs\end{tabular}} &
  \multicolumn{1}{c|}{0.8165} &
  0.6337 &
  0.5652 &
  0.5885 &
  0.7251 \\
SeQwen &
  \multicolumn{1}{l|}{\begin{tabular}[c]{@{}l@{}}Qwen2.5 7B fine-tuned using sequential learning approach\\ for a total of 8 epochs (3 epochs of classification followed by\\ 5 epochs of classification + explanation generation)\end{tabular}} &
  \multicolumn{1}{c|}{0.8283} &
  0.7253 &
  0.6763 &
  0.6911 &
  0.7768 \\ \hline
\end{tabular}%
}
\caption{Performance on the test set for Financial Misinformation Detection}
\label{tab:testFMD}
\end{table*}

\section{Results}
During the model selection phase, various models were assessed for both classification and joint classification \(+\) explanation generation on the development set to identify the top-performing models. For the classification task (Table \ref{tab:class}), Qwen2.5 7B delivered the strongest performance with micro F1 of \(0.8455\). Mistral 7B (micro F1 of \(0.8234\)) and Llama3 8B (micro F1 of \(0.8190\)) also performed admirably, demonstrating the ability of LLMs to detect misinformation in financial domains.\\
When models were fine-tuned for simultaneous classification and explanation generation, the performance declined slightly in terms of micro F1 score compared to classification-only fine-tuning, as shown in Table \ref{tab:class} and Table \ref{tab:devFMD}. This tradeoff highlights the challenge of optimizing for both tasks simultaneously. For instance, Qwen2.5 7B achieved a Micro F1 score of \(0.8322\) during joint fine-tuning, compared to \(0.8455\) in classification-only training, representing a small drop of 1.6\%. This shows Qwen's effectiveness in financial domains for interpretable misinformation detection. Mistral also performed admirably with ROUGE-1 of \(0.6710\), however, it lagged behind Qwen2.5 in the micro F1 score. These results highlight the strength of smaller, fine-tuned models like Qwen2.5 7B, which emerged as a clear leader in both classification and explanation tasks during the model selection phase.\\
Qwen2.5 7B was then fine-tuned using a sequential learning approach, termed SeQwen, which involved 3 epochs of classification-only fine-tuning followed by 5 epochs of joint fine-tuning for both classification and explanation generation. The performance improvements achieved using this approach are shown in Table \ref{tab:devFMD}. SeQwen outperformed its single-phase training counterparts, achieving a Micro F1 score of \(0.8366\), ROUGE-1 of \(0.7170\), ROUGE-2 of \(0.6639\), and ROUGE-L of \(0.6772\). Compared to Qwen2.5 7B fine-tuned for 5 epochs of joint training, SeQwen demonstrated improvements in all metrics, highlighting the advantages of staged, task-specific training.\\
To ensure a fair comparison, Qwen2.5 7B was also fine-tuned for a total of 8 epochs in a single-phase joint classification \(+\) explanation generation setup. Interestingly, while Qwen2.5 7B trained for 8 epochs (denoted as Qwen2.5 7B 8ep) achieved a slightly higher overall score than the 5-epoch counterpart (from \(0.7516\) to \(0.7552\) on the dev set), it still fell short of the performance achieved by SeQwen. This demonstrates that while extending training can offer marginal gains, the sequential learning strategy employed by SeQwen brings a more pronounced improvement across metrics, particularly in explanation quality as measured by ROUGE metrics.\\
This was further validated on the test set, as shown in Table \ref{tab:testFMD}. Compared to Qwen2.5 7B fine-tuned for 5 epochs of joint classification and explanation generation, SeQwen achieved improvements across all metrics, with the Micro F1 score increasing from \(0.8165\) to \(0.8283\), representing a \(1.4\)\% relative gain. For explanation generation, notable progress was seen in the ROUGE metrics: ROUGE-1 rose from \(0.6337\) to \(0.7253\) (a \(14.5\)\% increase), ROUGE-2 increased from \(0.5652\) to \(0.6763\) (\(19.7\)\% gain), and ROUGE-L improved from \(0.5885\) to \(0.6911\) (\(17.4\)\% increase). Additionally, the overall score improved from \(0.7251\) to \(0.7768\), reflecting a \(7.1\)\% improvement, emphasizing the synergistic effect of sequential fine-tuning in optimizing both classification and explanation generation.

\section{Conclusion}
Our results demonstrate the effectiveness of leveraging sequential fine-tuning approaches to address the dual challenges of misinformation detection and explanation generation in financial content. By first fine-tuning models like Qwen2.5 7B for classification and subsequently adapting them to generate explanations, we achieved significant performance improvements in both tasks. This progressive strategy allowed the model to specialize in identifying fraudulent content before learning to articulate clear, concise, and contextually relevant explanations, ensuring a robust balance between predictive accuracy and interpretability.\\
The findings underscore the importance of task-specific adaptation in large language models, particularly in complex domains such as finance, where both classification accuracy and transparency are critical. The superior performance of the SeQwen model highlights the potential of smaller, efficiently trained models when combined with tailored training strategies. This work establishes a foundation for building interpretable, domain-specific AI systems that not only detect misinformation but also enhance user trust through actionable insights and explainability. Future directions include exploring more advanced fine-tuning techniques and ensembling strategies to further enhance robustness and scalability in high-stakes applications.

\section*{Limitations}
While our approach demonstrated promising results, there are notable limitations that should be addressed in future work. First, the sequential fine-tuning strategy, while effective, requires careful balancing of training epochs for each stage to avoid catastrophic forgetting or overfitting, particularly for smaller datasets. Fine-tuning large language models such as Qwen2.5 7B and Llama3 8B demands substantial computational resources, which may limit accessibility for users with restricted hardware or budget. The models were fine-tuned in 4-bit precision due to computational limitations, and they may perform better in full-precision mode.\\
Additionally, the models' reliance on pre-existing knowledge embedded in their pre-trained weights may limit their ability to detect novel or domain-specific misinformation not covered during fine-tuning. Although our approach incorporates explanation generation to enhance interpretability, the quality and comprehensiveness of these explanations can still fall short in scenarios involving highly nuanced or ambiguous financial content. The ROUGE scores, while indicative of performance, may not fully capture the depth and correctness of explanations, necessitating further evaluation through human-in-the-loop methods.\\
Finally, the models were evaluated primarily on benchmark datasets, which, while reflective of real-world financial misinformation, may not account for rapidly evolving language trends or manipulation tactics in the financial domain. Future work should explore continual learning techniques and more dynamic datasets to address these challenges.
\bibliography{custom}

\appendix
\section{Appendix}
\label{sec:appendix}

\subsection{Confusion Matrix}
We provide the confusion matrix for classification performance of all the models we tested below:

\begin{figure}[H]
    \centering
    \includegraphics[width=0.9\columnwidth]{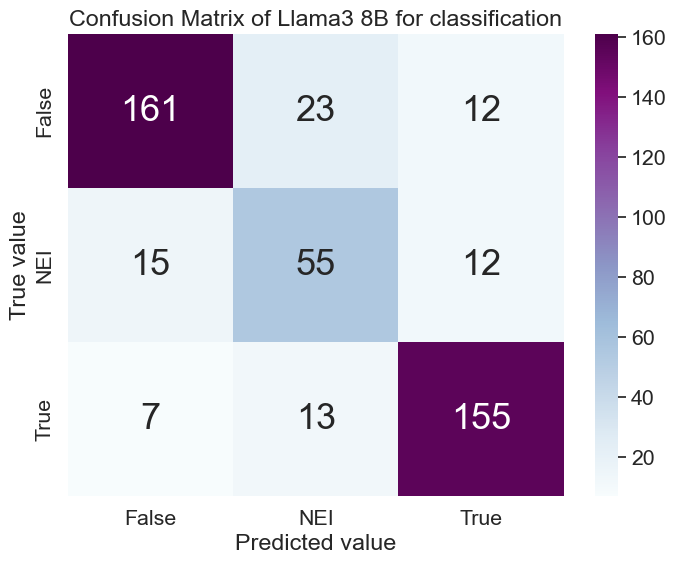}
    \caption{Llama3 8B's Confusion Matrix for classification on the dev set}
    \label{fig:llama_matrix}
\end{figure}

\begin{figure}[H]
    \centering
    \includegraphics[width=0.9\columnwidth]{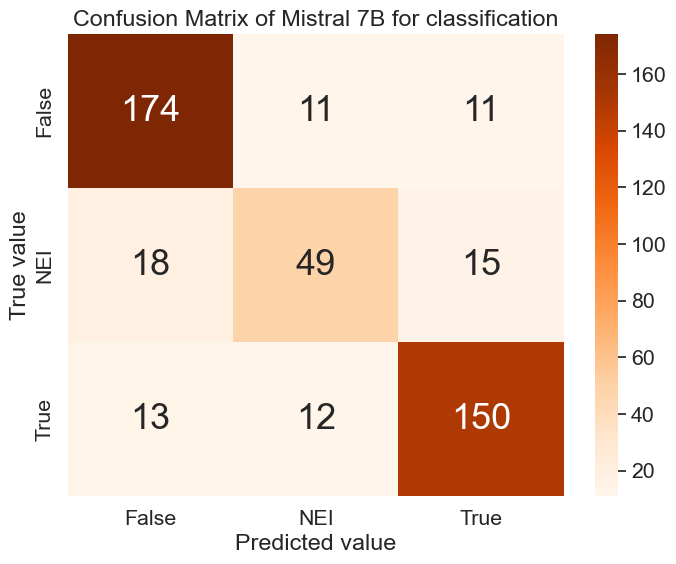}
    \caption{Mistral 7B's Confusion Matrix for classification on the dev set}
    \label{fig:mistral_matrix}
\end{figure}

\begin{figure}[H]
    \centering
    \includegraphics[width=0.9\columnwidth]{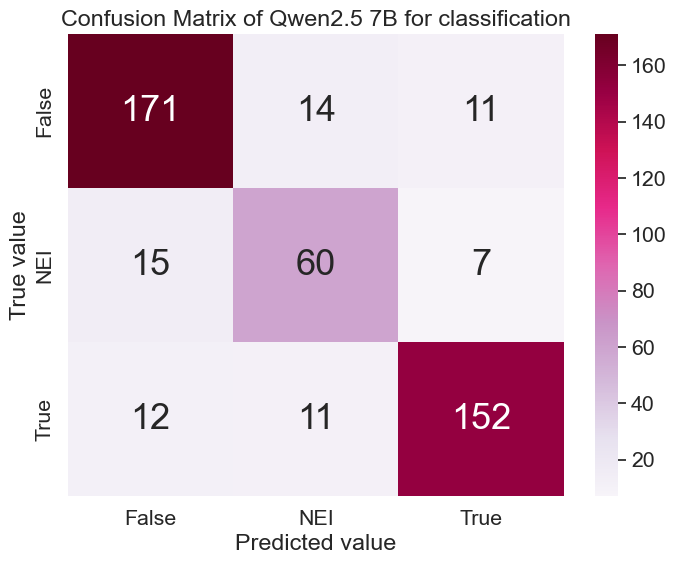}
    \caption{Qwen2.5 7B's Confusion Matrix for classification on the dev set}
    \label{fig:qwen7_matrix}
\end{figure}

\begin{figure}[H]
    \centering
    \includegraphics[width=0.9\columnwidth]{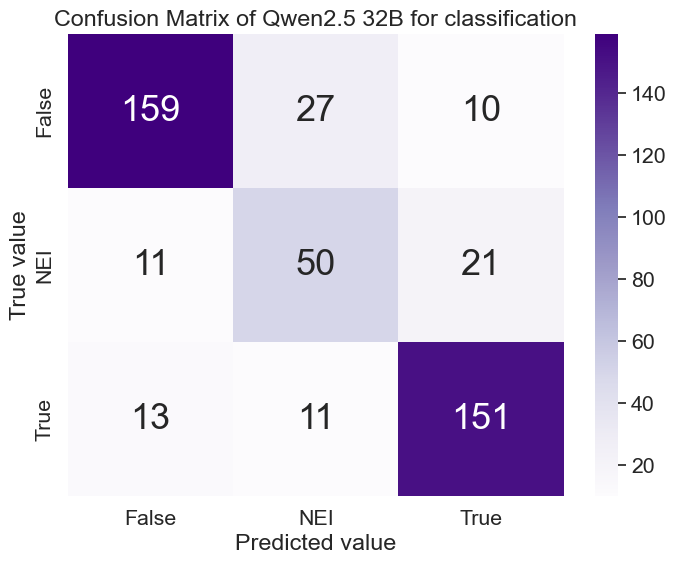}
    \caption{Qwen2.5 32B's Confusion Matrix for classification on the dev set}
    \label{fig:qwen32_matrix}
\end{figure}

\begin{figure}[H]
    \centering
    \includegraphics[width=0.9\columnwidth]{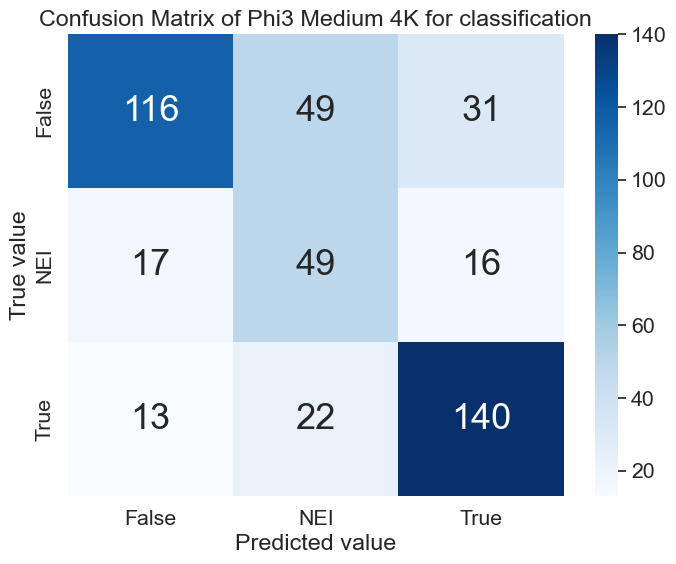}
    \caption{Phi3 Medium 4K's Confusion Matrix for classification on the dev set}
    \label{fig:phi_matrix}
\end{figure}

\begin{figure}[H]
    \centering
    \includegraphics[width=0.9\columnwidth]{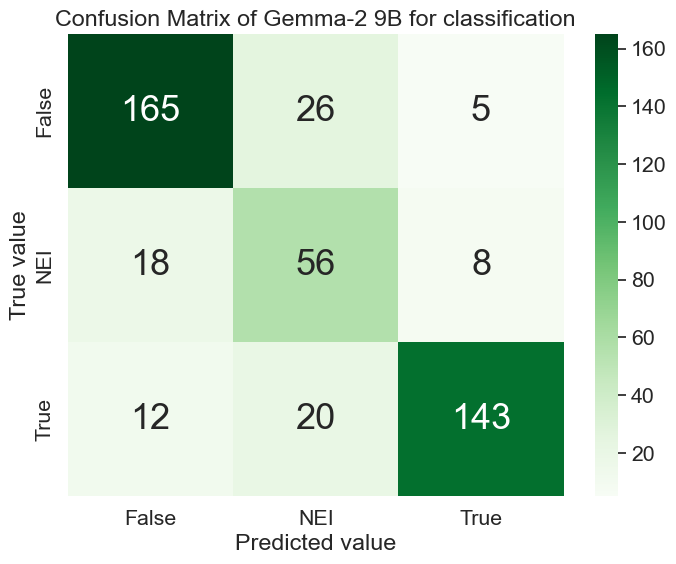}
    \caption{Gemma-2 9B's Confusion Matrix for classification on the dev set}
    \label{fig:gemma_matrix}
\end{figure}

\end{document}